\documentclass[conference]{IEEEtran}
\IEEEoverridecommandlockouts
\usepackage{cite}
\usepackage{amsmath,amssymb,amsfonts}
\usepackage{algorithmic}
\usepackage{graphicx}
\usepackage{textcomp}
\usepackage{xcolor}
\usepackage{float} 
\usepackage{hyperref} 
\usepackage{comment} 
\usepackage{caption} 
\usepackage{subcaption} 

\def\BibTeX{{\rm B\kern-.05em{\sc i\kern-.025em b}\kern-.08em
    T\kern-.1667em\lower.7ex\hbox{E}\kern-.125emX}}
\begin{document}

\makeatletter
\newcommand{\linebreakand}{
    \end{@IEEEauthorhalign}
    \hfill\mbox{}\par
    \mbox{}\hfill\begin{@IEEEauthorhalign}
}
\makeatother
\title{\vspace{18pt}
Learning to Find Missing Video Frames with Synthetic Data Augmentation: \\A General Framework and Application in Generating Thermal Images Using RGB Cameras
}

\author{

\IEEEauthorblockN{Mathias Viborg Andersen\thanks{M. Andersen, R. Greer, and M. Trivedi are with the Laboratory for Intelligent and Safe Automobiles. A. Møgelmose is with the Visual Analysis and Perception Lab. }}
\IEEEauthorblockA{ 
\textit{Laboratory for Intelligent \& Safe Automobiles (LISA)}\\
University of California San Diego \\
mvan19@student.aau.dk}
\and
\IEEEauthorblockN{Ross Greer}  
\IEEEauthorblockA{
\textit{Laboratory for Intelligent \& Safe Automobiles (LISA)}\\
University of California San Diego \\
regreer@ucsd.edu}
\linebreakand
\hspace{35pt}\IEEEauthorblockN{Andreas Møgelmose}
\IEEEauthorblockA{
\hspace{35pt}\textit{Visual Analysis and Perception Lab}\\
\hspace{35pt}Aalborg Universitet \\
\hspace{35pt}anmo@create.aau.dk}
\and
\hspace{35pt}\IEEEauthorblockN{Mohan M. Trivedi}
\IEEEauthorblockA{
\hspace{35pt}\textit{Laboratory for Intelligent \& Safe Automobiles (LISA)}\\
\hspace{35pt}University of California San Diego \\
\hspace{35pt}mtrivedi@ucsd.edu}
}

\maketitle

\begin{abstract}
Advanced Driver Assistance Systems (ADAS) in intelligent vehicles rely on accurate driver perception within the vehicle cabin, often leveraging a combination of sensing modalities. However, these modalities operate at varying rates, posing challenges for real-time, comprehensive driver state monitoring. This paper addresses the issue of missing data due to sensor frame rate mismatches, introducing a generative model approach to create synthetic yet realistic thermal imagery. We propose using conditional generative adversarial networks (cGANs), specifically comparing the pix2pix and CycleGAN architectures. Experimental results demonstrate that pix2pix outperforms CycleGAN, and utilizing multi-view input styles, especially stacked views, enhances the accuracy of thermal image generation. Moreover, the study evaluates the model's generalizability across different subjects, revealing the importance of individualized training for optimal performance. The findings suggest the potential of generative models in addressing missing frames, advancing driver state monitoring for intelligent vehicles, and underscoring the need for continued research in model generalization and customization.
\end{abstract}


\section{Introduction}

\begin{figure*}
    \centering
    \includegraphics[width=.98\textwidth]{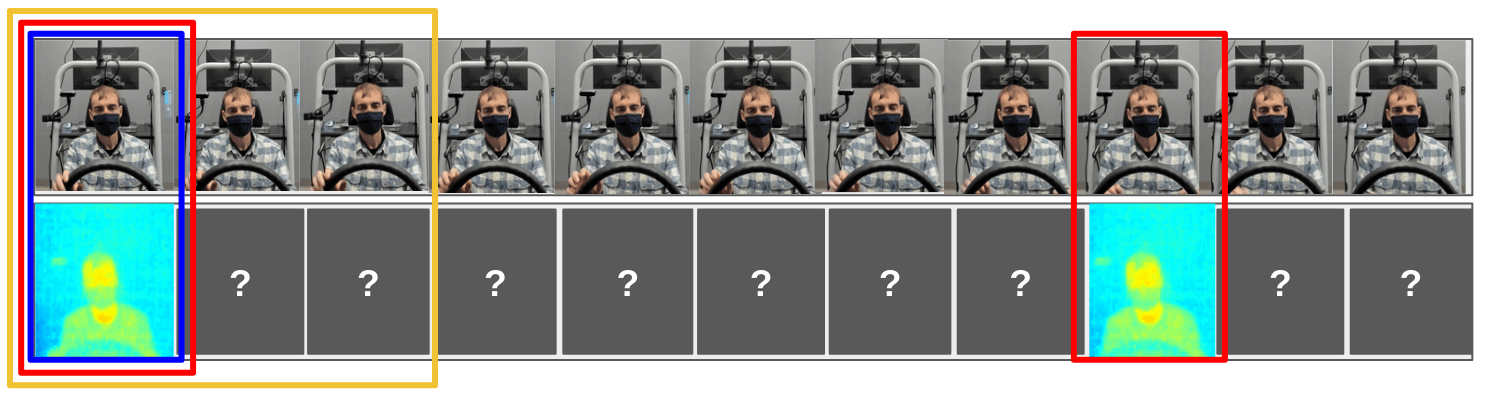}
    \caption{Many perspectives and modalities of data may contribute to robust driver state monitoring. Differing frame rates of sensors lead to an unavailability of ``complete" sets of data from all modalities for a given instance. Because many driver states are best inferred from temporal patterns, an ideal data stream would have constant availability of all sources at each instance. Without such a stream, models may be limited to instance inference (blue), complete-but-temporally-distant sequences (red), or incomplete-but-temporally-local sequences (yellow). By generating missing data, we can provide synthetic but useful representations to fill in these gray gaps, enabling accurate downstream state estimation models using pseudo-complete, temporally-local sequences.}
    \label{fig:framerate}
\end{figure*}
Many advanced driver assistance systems (ADAS) rely on accurate perception of the human driver by looking inside the intelligent vehicle cabin \cite{vora2018driver, ohn2015surveillance}. No single-view or sensing modality perfectly suits every task or design specification, so often a combination of views can be leveraged for enhanced driver state understanding \cite{greer2023multi}. However, different sensing modalities operate at different rates and often perform similarly to RGB using neural networks \cite{kantas2023raw}. For example, thermal cameras often operate at a fraction of the rate of cameras on the visible light spectrum, but are useful toward a variety of tasks in driver state monitoring by providing useful information for understanding occlusion, circulation, respiration, and other heat-related observations \cite{kajiwara2021driver, bole2016driver, mattioli2023thermal, kiashari2018monitoring, weiss2022head, palmero2016multi}. 

In many cases, it is useful for models to make observations based on a sequence of observations, rather than a single instance in time; doing so allows the modeling of dynamic events and driver actions \cite{rangesh2021autonomous, rangesh2021predicting, greer2023safe}, as well as inference reinforcement by repeated agreement between information observed at nearby times \cite{greer2023safesmart}. Considering again the multi-modal nature of driver observation systems, misaligned sensor rates are not a problem when data from each modality is considered individually, but requires careful synchronization and creative approaches to modeling in cases when data from one sensor may not be provided at the ideal rate for best utilizing the remaining sensors. We illustrate this problem, and associated problems and trade-offs, in Figure \ref{fig:framerate}; essentially, one can either use a high-frequency model with intermittent missing data from low-frequency sensors, or a low-frequency model which may sacrifice temporal precision of inference. The risk of using low-frequency sensors is the mismatch that occurs when the driver takes an action which changes state in a way that can be observed by one sensor but missed by another; this can create conflicting information and multimodal model confusion if not handled properly, illustrated in Figure \ref{fig:mismatch}. 

\begin{figure}
    \centering
    \includegraphics[width=.205\textwidth]{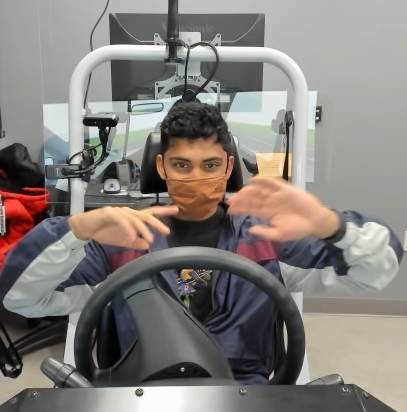}
    \includegraphics[width=.24\textwidth]{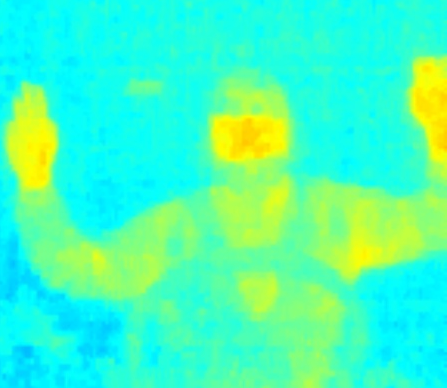}
    \caption{When sensors operate at different rates, it is possible that the temporally-nearest measurement to a given instance may have taken place before a significant action for one sensor, and after the action for another. In the above example, the driver has abruptly moved his hands closer to the wheel; however, the thermal camera has not yet processed another signal to capture this motion. So, if both ``most recent" signals are sent to a multimodal model meant to estimate a driver's takeover readiness (e.g. proximity of hands to the steering wheel), the model would have a large amount of uncertainty from modal disagreement.}
    \label{fig:mismatch}
\end{figure}

In this research, we propose a solution using a generative model to create pseudo-complete data samples, pairing both real (low frequency) thermal imagery and generated samples with high-frequency visible light images captured from multiple perspectives. 

\section{Related Research}


\subsection{Synthetic Thermal Images for Data Augmentation}

Various modes of image data, from RAW \cite{kantas2023raw} to RGB to IR, and even to non-light-based imagery such as temperature, provide utility in a variety of applications; further, cross-modality image synthesis is useful as a data augmentation strategy in a variety of tasks, such as terrain classification (visible to IR) \cite{iwashita2019mu}, tissue segmentation (MRI to CT) \cite{chen2021diverse}, and heart observation (various CMR medical imaging techniques) \cite{wang2022cross}. Specific to human interaction, Hermosilla et al. \cite{hermosilla2021thermal} generate thermal facial images from noise using StyleGAN2, in order to build a robust dataset with tunable features that can be used to train deep learning models to detect faces within thermal images, a task which is generally more difficult than detection from visible light images but may be useful in certain sensing environments. Our research similarly utilizes GAN as the underlying generative learning paradigm, with the common goal of augmenting datasets towards problems in human subject feature extraction and understanding.

\subsection{Translating between Thermal and Visible Light}

Deep learning has revolutionized image processing and opened up new possibilities for image generation. Deep learning models, particularly generative adversarial networks (GANs), have demonstrated remarkable capabilities in translating images between different domains. Relevant to our domains of interest (visible light and thermal signature), Abdrakhmanova et al. \cite{abdrakhmanova2021speakingfaces} create the SpeakingFaces dataset and explore models to generate visible images from thermal images, since facial features for landmark recognition and detection are too obscure in thermal imagery, reducing possible use cases in HCI, biometric authentication, and other systems. They use CycleGAN and CUT models to map thermal face images to the visual spectrum, evaluating both the generated Fréchet inception distance as well as the ability of the generated models to produce the correct output on downstream facial landmark detection tasks.

Li et al. \cite{li2023feasibility} introduced the dual-attention generative adversarial network (DAGAN) for the  generation of thermal images from visible light, but outside the realm of human subjects and in the domain of fire safety, useful in estimating locations and temperatures of flames in room fires. Though the temperature range, contrast, and locality on human subjects is significantly different from that of an open flame, their research provides a strong indication that GAN architectures are suited to the task.


\subsection{Predicting Missing Frames}
For real-world data, the expected translational motion of objects between frames has allowed for missing individual frames to be estimated using a Kalman filtering approach \cite{chaubey2023estimation}, and for larger segments of up to 14 frames to be created from a fully convolutional model \cite{li2019here}. However, such methods which effectively ``in-paint" or ``in-between" video sequences are ineffective toward replacing larger missing sequences (in this case, approximately 1 frame available in every 5-frame period). Further exacerbating this issue, the frame rate is slow enough that basic human motion changes state at a scale faster than the sensor records-- meaning that even knowing the surrounding frames may be insufficient to fill in the activity of the frames in between. Fortunately, in this problem setting, we have at our disposal additional information from the ``missing" times, rather than just its surrounding pieces.

\section{Methods}

\subsection{Synthetic Data Augmentation}

In the case presented in this paper, we consider a downstream goal task of driver state estimation, and we have a set of thermal and visible light images which are used to supervise the training of this task. However, we also have a set of visible light images with no matching thermal imagery; by training an intermediate model which generates associated thermal images, and then using these generated images in the training of a driver state estimation model, we can augment the dataset to enable the use of models which operate over complete sets of sensor data at a high frequency.

\subsection{Algorithm}

We train and evaluate two conditional generative adversarial network (cGAN) architectures \cite{mirza2014conditional}; the pix2pix architecture \cite{isola2018imagetoimage} and the CycleGAN architecture \cite{zhu2017unpaired}. These models consist of two competing neural networks: a generator that produces thermal images from RGB inputs and a discriminator that distinguishes between real thermal images and generated ones. Through adversarial training, the generator learns to synthesize realistic thermal images that are indistinguishable from real ones. The cGAN architecture of pix2pix is seen in Figure \ref{fig:pix2pixFlow}, and a demonstration of its iterative learning is shown in Figure \ref{fig:genflow}.

\begin{figure}
    \centering
    \includegraphics[width=1\linewidth]{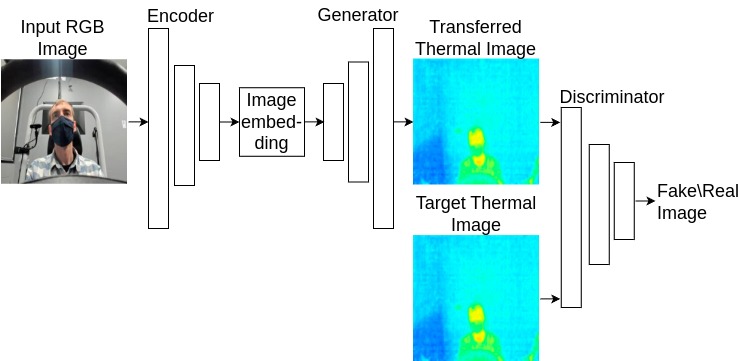}
    \caption{The flow of pix2pix applied in this work.}
    \label{fig:pix2pixFlow}
\end{figure}

Our output for all methods is a thermal image, like the example shown in Figure \ref{fig:datasetTypes}a. We evaluate four types of input for experimental comparison:
\begin{enumerate}
    \item Front-View RGB (Figure \ref{fig:datasetTypes}b)
    \item Single-Subject Four-View RGB, tesselated (Figure \ref{fig:datasetTypes}c)
    \item Single-Subject Four-View RGB, stacked (Figure \ref{fig:datasetTypes}d)
\end{enumerate}
Though these input styles may appear unusual, their compactness allows for training on a single GPU system, and on an efficient model architecture, as opposed to creating three additional convolutional ``heads" to extract features from the individual images. Though there are some false edges introduced due to the stacking structures, our results show that these collage images are still able to significantly outperform single-view learning. Additionally, we create one further experiment to evaluate for the effects of single-subject training versus multi-subject training for the Front View. 


\subsection{Dataset}
The dataset utilized in this study is notably comprehensive, incorporating various perspectives beneficial for developing and evaluating innovative approaches to generating thermal images from RGB inputs as well as driver state monitoring in general. It consists of distinct viewpoints, including \textit{thermal}, \textit{front}, \textit{overhead}, \textit{profile}, and \textit{tablet} orientations, as displayed in Figure \ref{fig:availableDataStructure}.
\begin{figure}[H]
    \centering
    \includegraphics[width=1\linewidth]{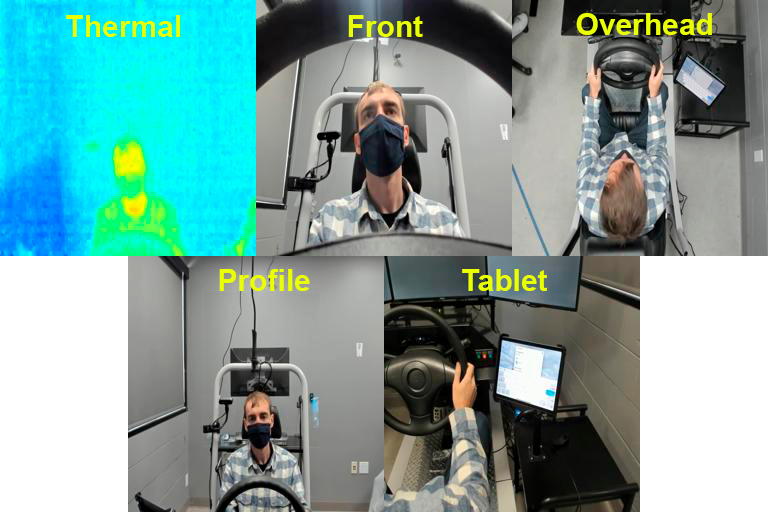}
    \caption{Example images showcasing perspectives captured from used cameras within our simulator setup. }
    \label{fig:availableDataStructure}
\end{figure}

The dataset consists of captures of 17 subjects seated at a simulated driver's seat. Notably, the RGB images are captured at a rate of approximately 30 frames per second (fps), while the thermal imaging data is sourced from a thermal camera operating at less than 9 fps, and represent a range of -20\textdegree C to 300\textdegree C, scaled to [0, 255]. Thorough synchronization and preparation procedures have been applied to ensure the optimal integration of the multi-view data.

From each subject we create a collection of 500 thermal + RGB image synchronous groups. In each experiment, an allocation of 80 \% for training, 10 \% for validation, and 10 \% for testing has been used. In the case of training on the aggregate pool of all subjects, we randomly select 5,000 of the 8,500 samples to account for our available training hardware.

\begin{figure}
    \centering
    \begin{minipage}[b]{0.45\linewidth}
        \centering
        \includegraphics[width=\linewidth]{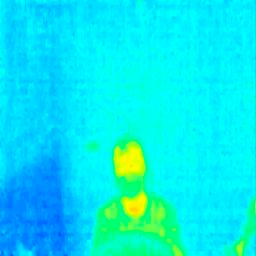}
        \caption*{(a) Thermal ground truth.}
    \end{minipage}
    \begin{minipage}[b]{0.45\linewidth}
        \centering
        \includegraphics[width=\linewidth]{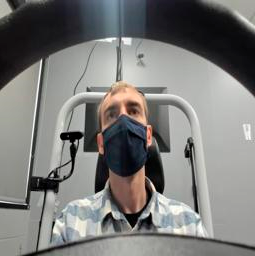}
        \caption*{(b) Front-View.}
    \end{minipage}
    \begin{minipage}[b]{0.45\linewidth}
        \centering
        \includegraphics[width=\linewidth]{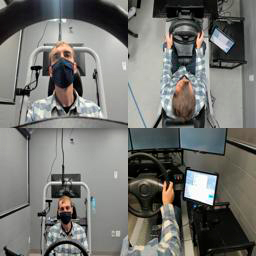}
        \caption*{(c) Four-View, Tessellated.}
    \end{minipage}
    \begin{minipage}[b]{0.45\linewidth}
        \centering
        \includegraphics[width=\linewidth]{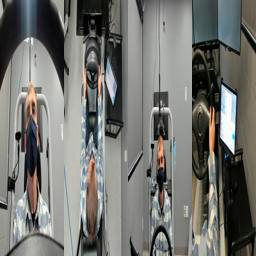}
        \caption*{(d) Four-View, Stacked.}
    \end{minipage}
    \caption{In our experiments, different inputs are evaluated on their potential for generating an image similar to the thermal ground truth.}
    \label{fig:datasetTypes}
\end{figure}

\begin{figure}
    \centering
    \includegraphics[trim=3.5cm 13.5cm 2.5cm 13.5cm, clip, width=0.49\textwidth]{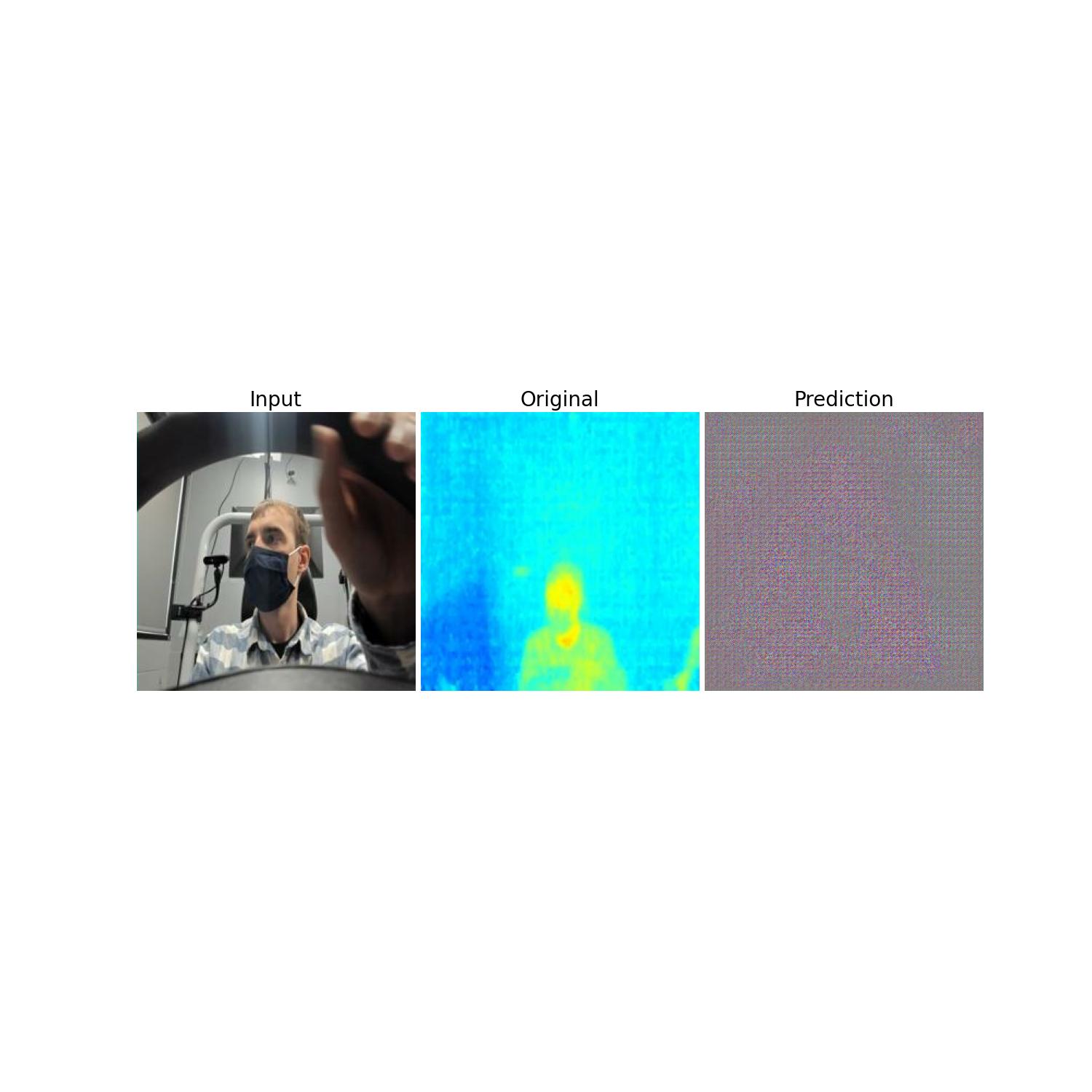}
    \includegraphics[trim=3.5cm 13.5cm 2.5cm 14.5cm, clip, width=0.49\textwidth]{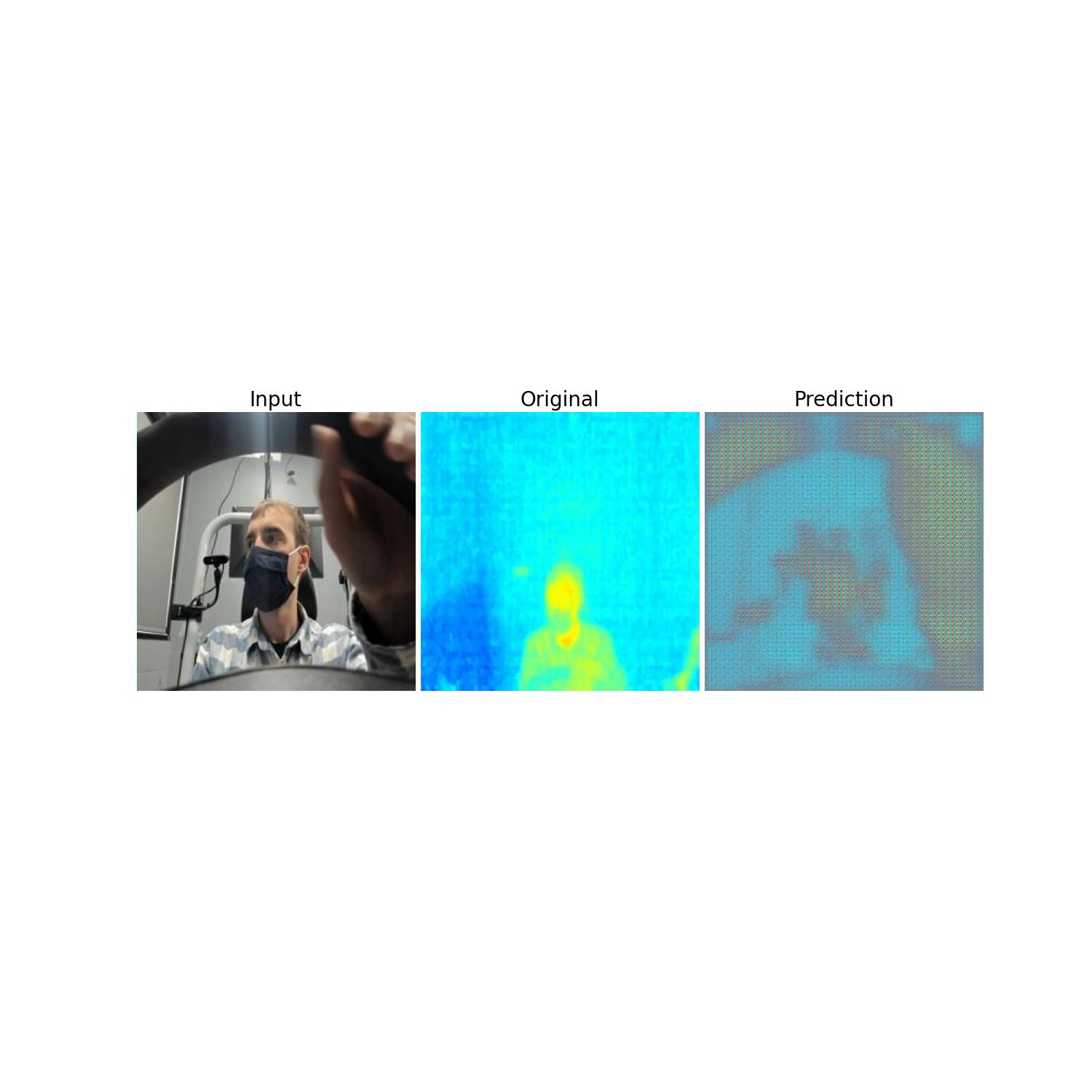}
    \includegraphics[trim=3.5cm 13.5cm 2.5cm 14.5cm, clip, width=0.49\textwidth]{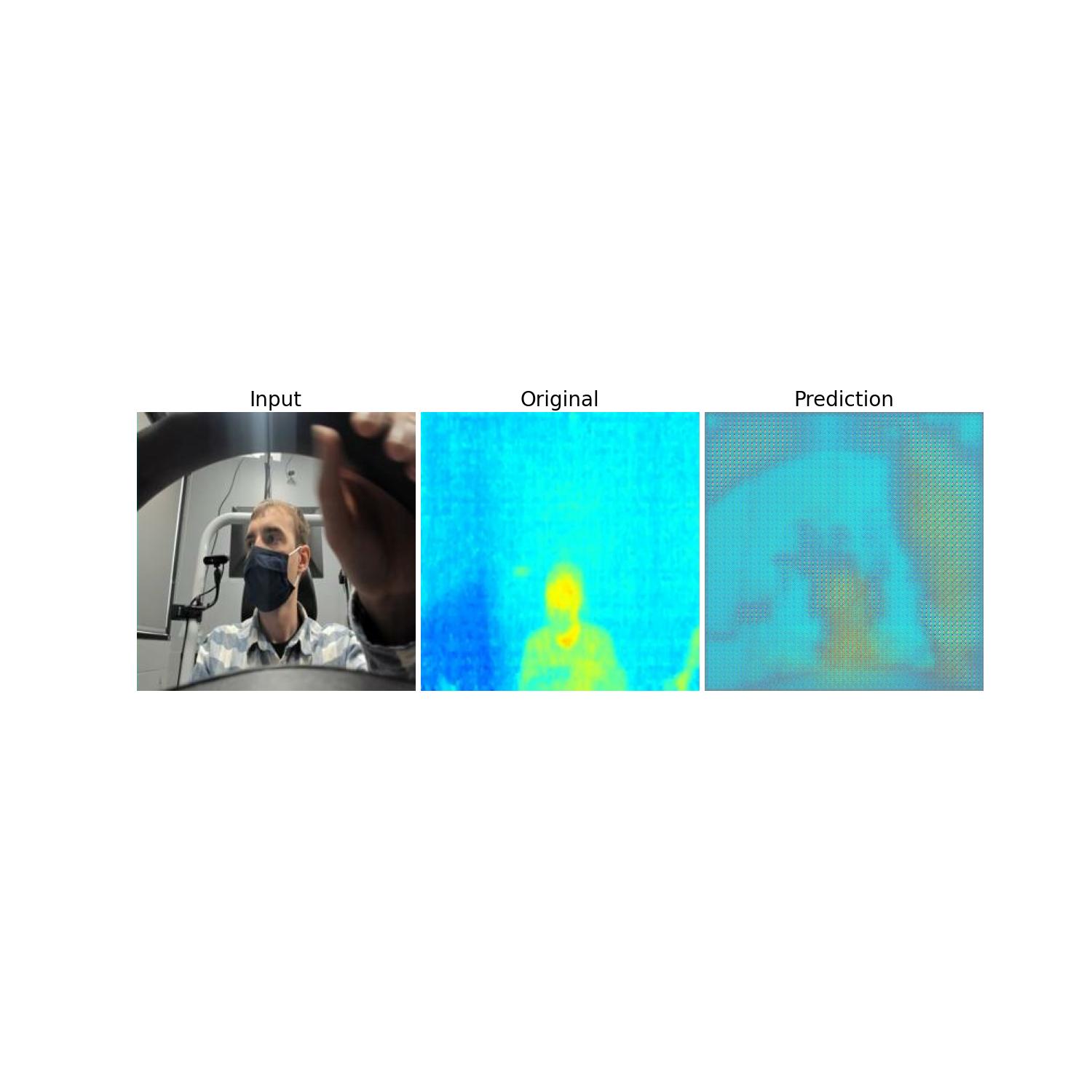}
    \includegraphics[trim=3.5cm 13.5cm 2.5cm 14.5cm, clip, width=0.49\textwidth]{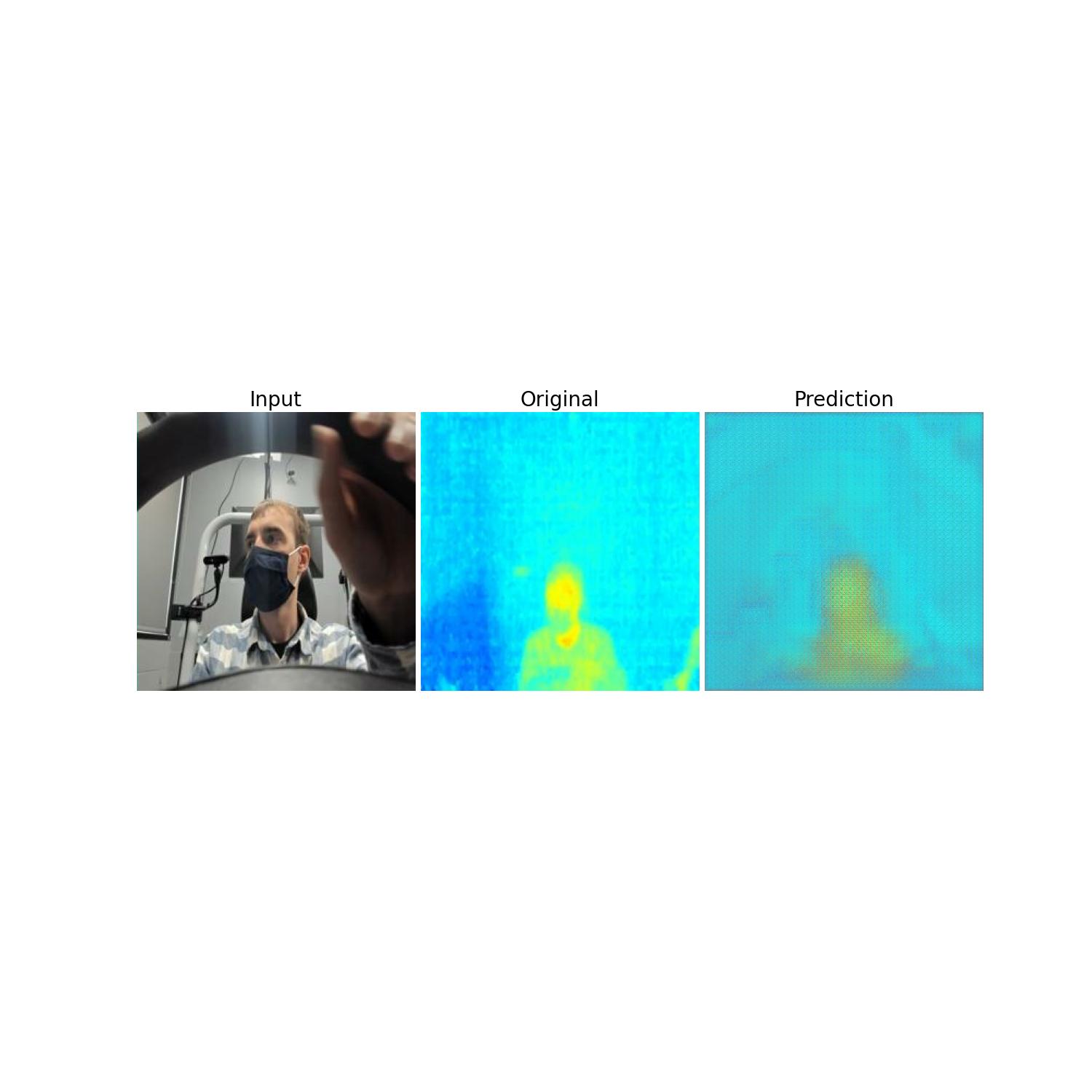}
    \includegraphics[trim=3.5cm 13.5cm 2.5cm 14.5cm, clip, width=0.49\textwidth]{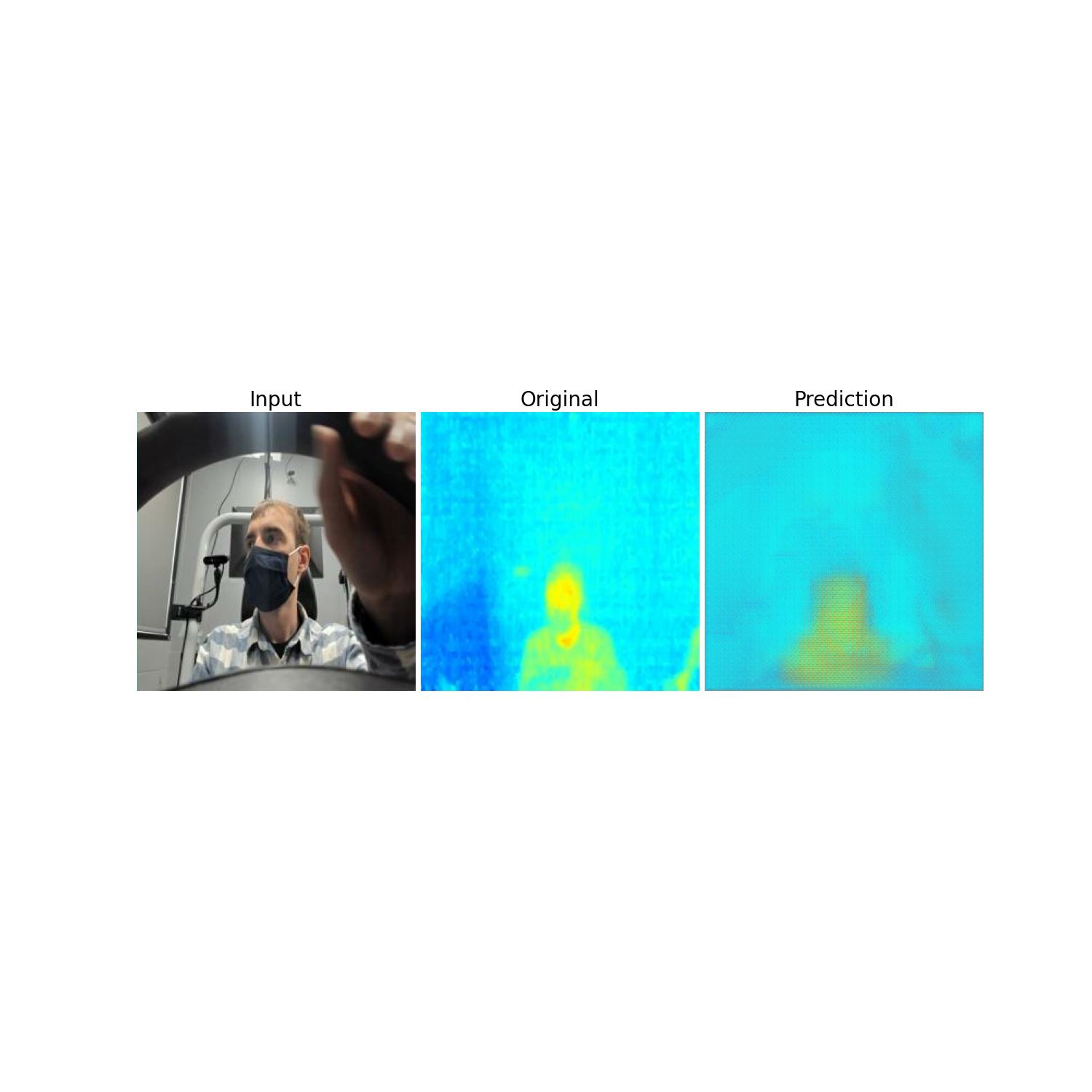}
    \includegraphics[trim=3.5cm 13.5cm 2.5cm 14.5cm, clip, width=0.49\textwidth]{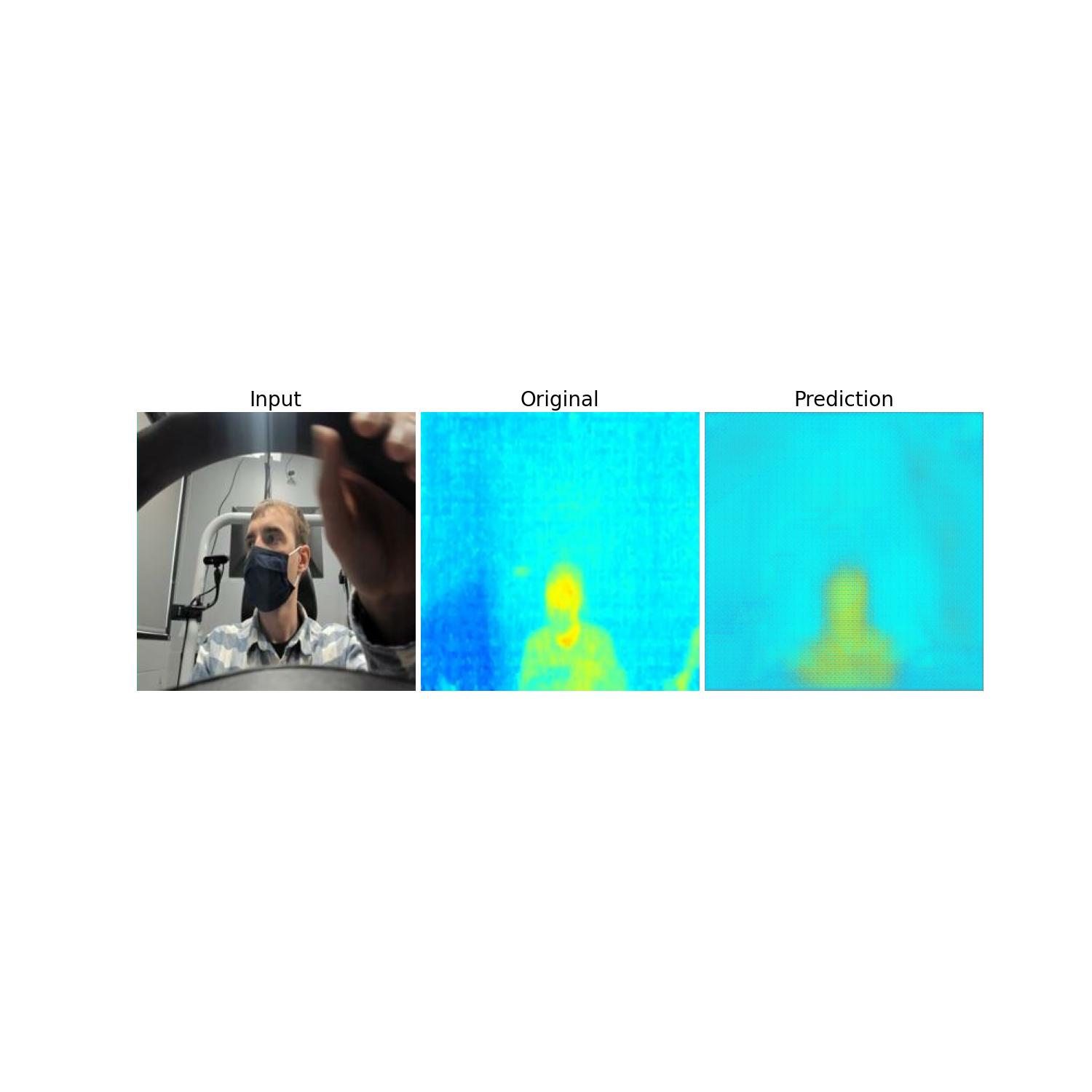}
    \includegraphics[trim=3.5cm 13.5cm 2.5cm 14.5cm, clip, width=0.49\textwidth]{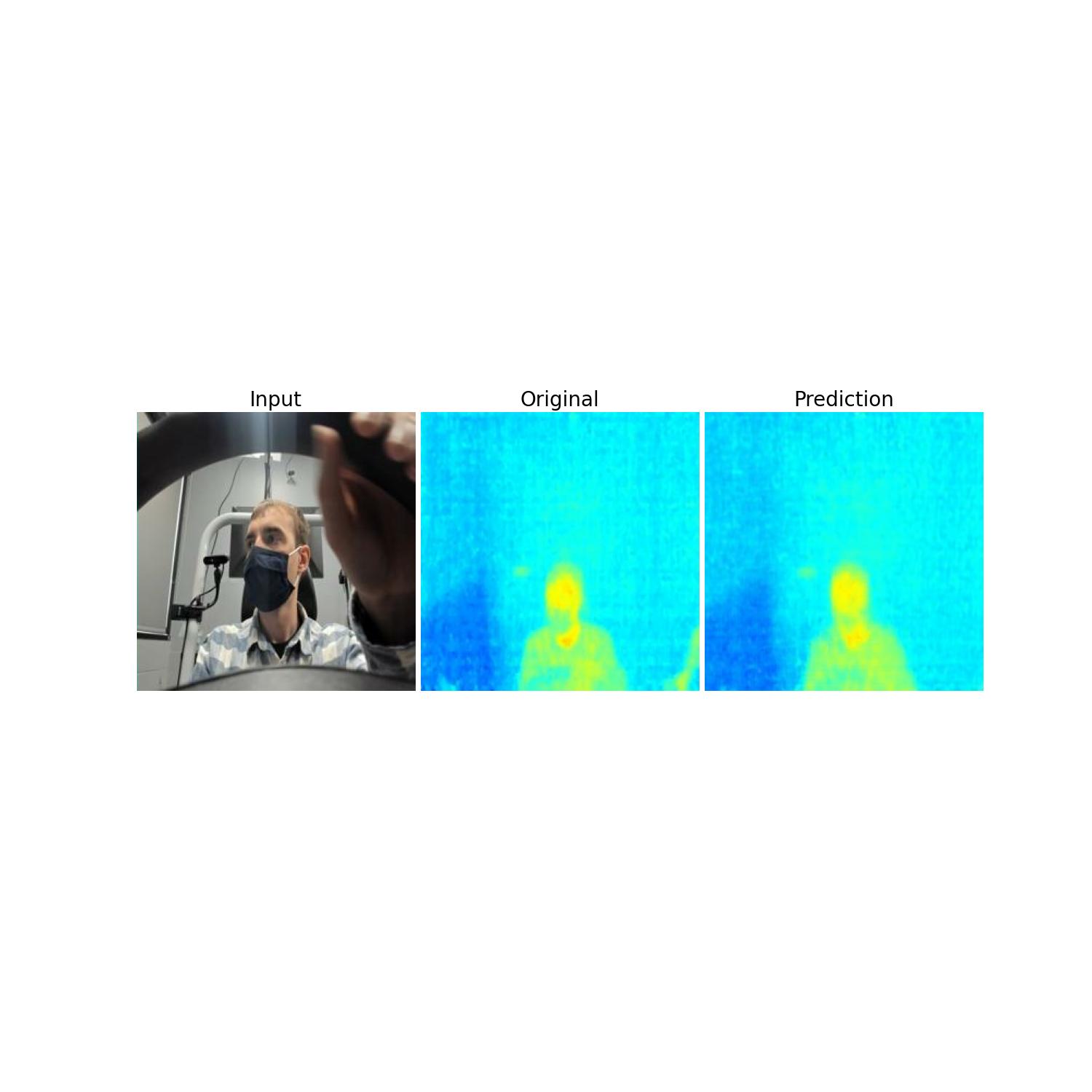}
    \caption{From top to bottom, we show the generator output at different iterations of training. Originally, the generator produces a random image, and refines its output to match the intended thermal image over time. These images are separated by only 10 iterations each, except for the final image which represents a jump to 20,000 iterations.}
    \label{fig:genflow}
\end{figure}

\section{Experimental Evaluation}
\subsection{Generative Architecture}
We first compare the performance of the pix2pix architecture versus the CycleGAN architecture, which has proven useful in past data augmentation for driver state monitoring \cite{rangesh2022gaze}. As shown in Table \ref{tablecycle}, the performance of the CycleGAN is significantly subpar to pix2pix. We expect that this is due to the attempt by CycleGAN to reconstruct the original input from its generated output during training; because the thermal image is significantly lossy compared to the amount of information in the visible light images, this relationship is more difficult to model beyond approximation. For interpretation, we note that all experimental error values are on normalized pixel values in the range [0, 1], meaning that most errors range around 5-6\% of the pixel range.

\begin{table}
    \centering
    \caption{Comparison of Model Architectures when training front-view.}
    \begin{tabular}{c|c|c}
        Method & Average Test L1 Error & Standard Deviation \\
        \hline 
        CycleGAN & 0.2179 & 0.0633 \\ \hline
        \textbf{pix2pix} & \textbf{0.0676} & \textbf{0.0106} \\
    \end{tabular}
    \label{tablecycle}
\end{table}

\subsection{Input Style}

The results of our experiments comparing the three different styles of input are presented in Table \ref{tab:results}. We find that a combination of views outperforms generation from a single-view, Figure \ref{fig:frontViewResults}, (perhaps assisting in understanding hand and posture positioning and respective heat signatures), and that the stacked-view and tesselated-view, Figures \ref{fig:sidebysideViewResults}\&\ref{fig:checkViewResults}, of the images provides an efficient and effective input, evidenced by the lowest average L1 error (0.0559) and a comparatively low standard deviation (0.0093). This suggests that considering spatial relationships by multi-view information enhances the model's accuracy in thermal image generation, as seen in Figure \ref{fig:sidebysideViewResults}.



\begin{table}
    \centering
    \caption{Comparison of Input Style; Average Performance Across 17 Subjects.}
    \begin{tabular}{c|c|c}
        \hline
        Dataset & Average Test L1 Error & Standard Deviation \\\hline \hline
        Front-View & 0.0676 & 0.0106 \\ \hline
        Four-View, Tessellated & 0.0587 & 0.0109 \\ \hline
        \textbf{Four-View, Stacked} & \textbf{0.0559} & \textbf{0.0093} \\ \hline
    \end{tabular}
    \label{tab:results}
\end{table}

\subsection{Subject Generalizability}
The results of our experiments on model generalizability to multi-subject training data is presented in Table \ref{tab:results_multi}. We find that though the training dataset size grows significantly (17$\times$) and still includes the original training data, the additional subjects seem to contribute more to model confusion than generalized pattern creation, showing a higher average L1 error of 0.1116 and supporting the idea that these models are best trained on an individual basis. The relatively weak performance when trained on the more diverse set of data can be observed in Figure \ref{fig:multifrontViewResults}.

\begin{table}
    \centering
    \caption{Comparison of Single vs. Multi-Subject Training on Front View.}
    \begin{tabular}{c|c|c}
        \hline
        Dataset & Average Test L1 Error & Std. Deviation \\\hline \hline
        \textbf{Single-Subject Training} & \textbf{0.0676} & \textbf{0.0106} \\ \hline
        Multi-Subject Training & 0.1116 &  0.0186\\ \hline
    \end{tabular}

    \label{tab:results_multi}
\end{table}


\section{Concluding Remarks}
Our study on generating thermal images from RGB counterparts highlights promising results with effective prediction approximation. However, the persisting challenge of the missing frames issue underscores the complexity of the task, demanding further research, as predictions must be near perfect. The stacked generation approach proved most successful, emphasizing the importance of spatial relationships. Despite these advancements, the model's generalization between subjects remains the worst performer, warranting continued efforts for improved adaptability across diverse scenarios so that singular models may be trained and deployed, and motivating research into the potential of small-data fine-tuning for model customization to individual drivers. The value of observing time-varying patterns is beneficial to many autonomous driving applications \cite{greer2024patterns, doshi2010examining, tawari2014robust}, providing a means to infer useful, high-frequency cues from temporal dynamics of the observed subject.  

In summary, our generative approach shows potential to address the missing frames problem (caused by sensor frame rate mismatches and intermittent failures), providing a means for higher frequency driver state monitoring for enhanced intelligent vehicle awareness and rapid, safe decision-making. 

\begin{figure}[H]
    \centering
    \includegraphics[width=1\linewidth]{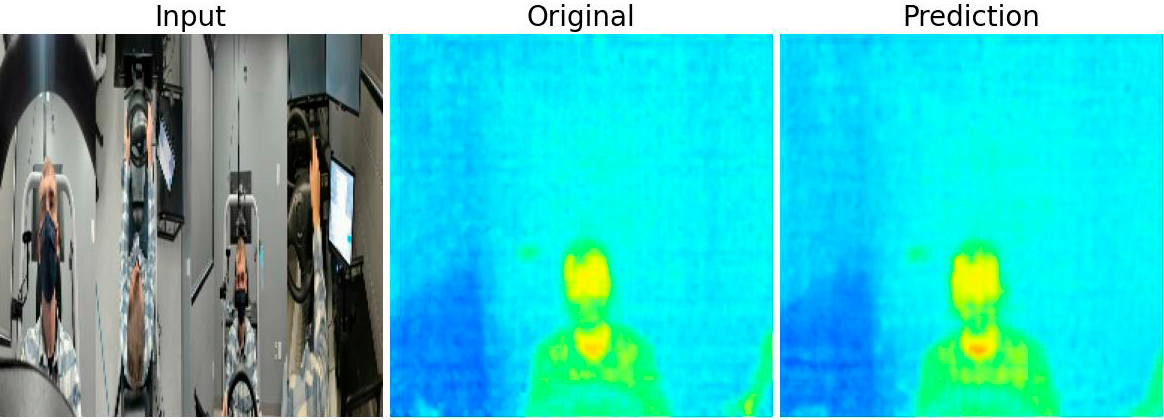}
    \caption{Single-Subject Four-View, Stacked Prediction Example.}
    \label{fig:sidebysideViewResults}
\end{figure}

\begin{figure}[H]
    \centering
    \includegraphics[width=1\linewidth]{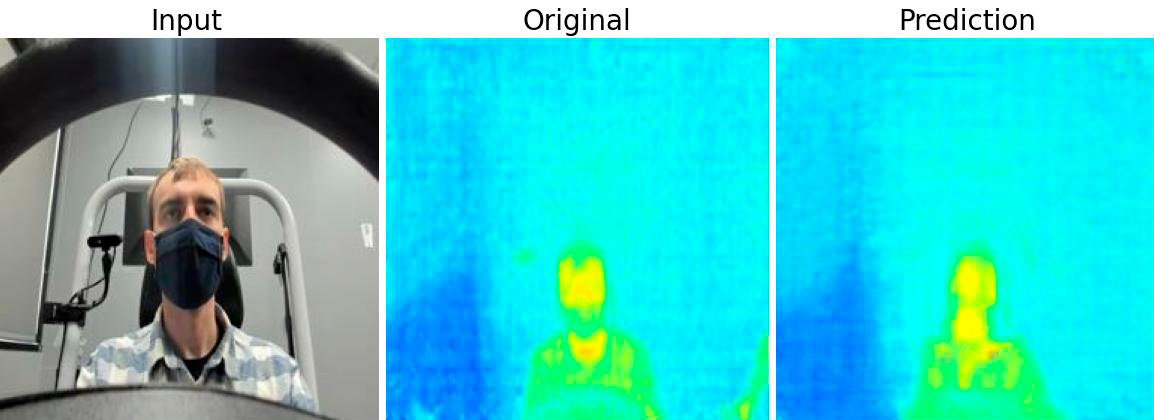}
    \caption{Multi-Subject Front-View Prediction Example.}
    \label{fig:multifrontViewResults}
\end{figure}


\begin{figure}[H]
    \centering
    \includegraphics[width=1\linewidth]{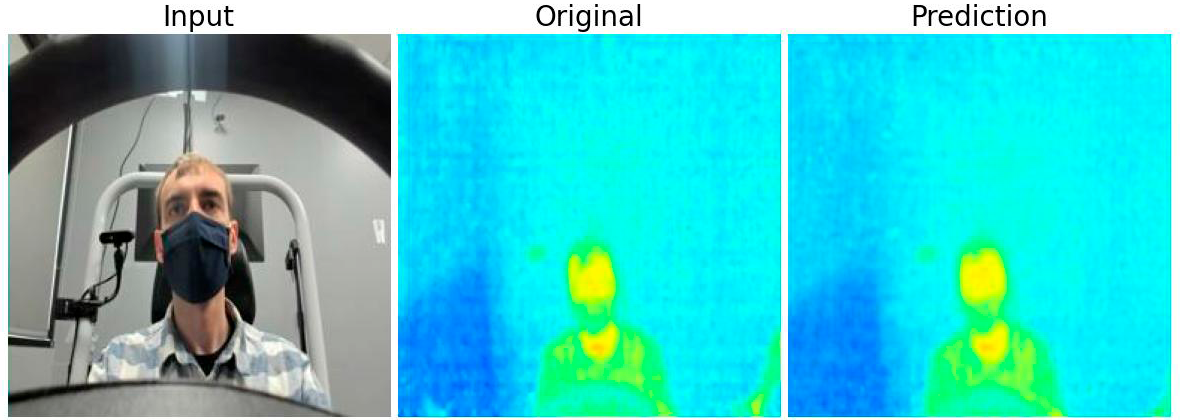}
    \caption{Single-Subject Front-View Prediction Example.}
    \label{fig:frontViewResults}
\end{figure}

\begin{figure}[H]
    \centering
    \includegraphics[width=1\linewidth]{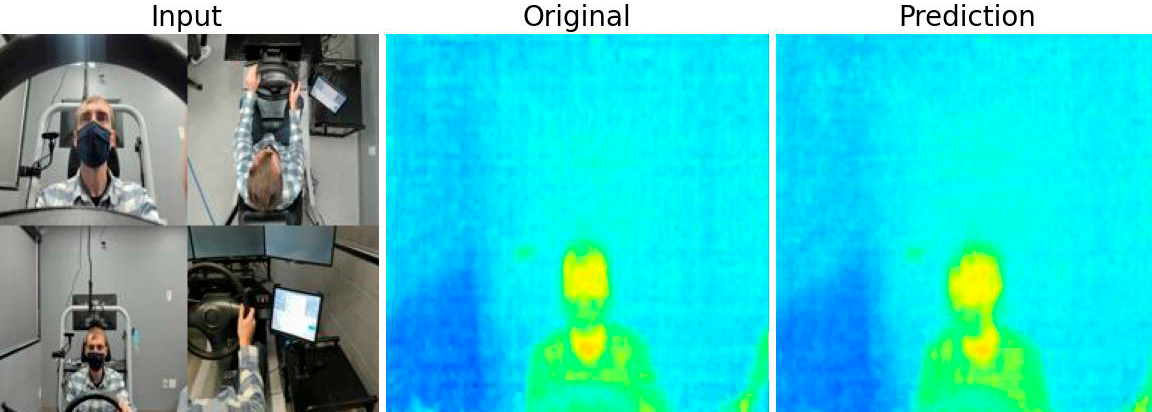}
    \caption{Four-View, Tessellated Prediction Example.}
    \label{fig:checkViewResults}
\end{figure}

\section*{Acknowledgements}

The authors would like to thank Sumega Mandadi for her assistance in the synchronization and management of the experimental dataset.

\bibliographystyle{ieeetr}
\bibliography{refs}


\end{document}